\documentclass[preprint,review,10pt]{elsarticle}

\usepackage{graphicx}
\usepackage{amsmath}
\usepackage{amssymb}
\usepackage{algorithmic}
\usepackage{algorithm}
\usepackage{subfigure}

\begin{document}

\begin{frontmatter}

\title{Regularized maximum correntropy machine}

\author[KAUST]{Jim Jing-Yan Wang}

\author[UTSA]{Yunji Wang}

\author[HKUST]{Bing-Yi Jing}

\author[KAUST]{Xin Gao \corref{cor1}}

\address[KAUST]{Computer, Electrical and Mathematical Sciences and Engineering Division, King Abdullah University of Science and Technology (KAUST), Thuwal 23955-6900, Saudi Arabia}

\address[UTSA]{Electrical and Computer Engineering Department, The University of Texas at San Antonio, San Antonio, TX 78249, USA}

\address[HKUST]{Department of Mathematics, Hong Kong University of Science and Technology, Kowloon, Hong Kong}

\cortext[cor1]{Correspondence should be addressed to Xin Gao. Tel: +966-12-8080323.}

\begin{abstract}
In this paper we investigate the usage of regularized correntropy framework for learning of
classifiers from noisy labels.
The class label predictors learned by minimizing transitional loss functions are sensitive to the
noisy and outlying labels of training samples, because the transitional loss functions are equally applied to all the samples. To solve this problem, we propose to learn the class label predictors by maximizing the correntropy between the predicted labels and
the true labels of the training samples, under the regularized Maximum Correntropy Criteria (MCC) framework.
Moreover, we regularize the predictor parameter to control the complexity of the
predictor.
The learning problem is formulated by an objective function considering the parameter regularization and
MCC simultaneously. By optimizing the objective function alternately, we develop a novel
predictor learning algorithm. The experiments on two challenging pattern classification tasks
show that it significantly outperforms the machines with transitional loss functions.
\end{abstract}


\end{frontmatter}

\section{Introduction}
\label{sec:intro}

The classification machine design has been a basic problem in the pattern recognition field.
It tries to learn an effective predictor to map the feature vector of a sample to its class label \cite{lu2015semantic,li2011k,lu2015noise,li2013online,zhou2014spatial,zhou2013adaptive,lu2013learning,zhou2010region,lu2013unified,zhou2013sub}.
We study the supervised multi-class learning problem with $L$ classes.
Suppose we have a training set denoted as
$\mathcal{D}=\{(x_i,y_i)\},i=1,\cdots,N$,
where $x_i =[x_{i1},\cdots, x_{iD}]^\top \in \mathbb{R}^D$ is the $D$
dimensional feature vector of the $i$-th training sample, and $y_i\in \{1,\cdots,L\}$ is the class label of
$i$-th training sample.
Moreover, we also denote the label indicator matrix
as $Y =[Y_{li}] \in \mathbb{R}^{L\times N}$, and $Y_{li} = 1$ if $y_i=l$, and $-1$ otherwise.
We try to learn $L$ class label predictors $\{f^l_\theta(x)\},l=1,\cdots,L$ for the multi-class learning
problem, where
$f^l_\theta(x)$ is the predictor for the $l$-th class and $\theta$
is its parameter.
Given a sample $x_i$, the output of the $l$-th predictor is denoted as $f^l_\theta(x_i)$,
and we further denote the prediction result matrix
as $F_\theta=[{F_\theta}_{li}]\in \mathbb{R}^{L\times N}$,
and ${F_\theta}_{li} = f^l_\theta(x_i)$.
To make the prediction as precise as possible,
the target of predictor learning is to learn  parameter $\theta$, so that
the difference between {true class labels of the training samples in $Y$ and the prediction results in $F_\theta$} could be minimized, while keeping the complexity of the predictor as low as possible.
To measure how well the prediction results fit the true class label indicator,
several loss functions $L(F_\theta, Y)$ could be considered to compare the prediction results in $F_\theta$ against the
true class labels of the training samples in $Y$, such as the 0-1 loss function, the square loss function, the hinge loss function, and the logistic loss function.
We summarize various loss functions in Table \ref{tab:loss}.

\begin{table}[!h]
\centering
\caption{Various empirical loss functions for predictor learning}
\label{tab:loss}
\begin{tabular}{p{0.15\textwidth}p{0.35\textwidth}p{0.5\textwidth}}
\hline
Title & Formula of $L(F_\theta, Y)$ & Notes \\
\hline\hline
0-1 Loss&
$\sum_{i,l} \mathbb{I}[{F_\theta}_{li} Y_{li}<0]$,
where $\mathbb{I}(\cdot) $ is the indicator function and
$\mathbb{I}(\cdot) =1$ if $(\cdot)$ is true, $0$ otherwise.
&
The 0-1 loss function is NP-hard to optimize, non-smooth and non-convex.\\
\hline
Square Loss&
$\sum_{i,l} [{F_\theta}_{li} - Y_{li}]^2=||F_\theta- Y||^2
$,
where $\circ$ denotes the element
wise product of two matrices,
and $\textbf{1}_{N\times L}$ is a $N\times L$ matrix with all elements of ones.
&
The square loss function is a convex upper bound on the 0-1 loss.
It is smooth and convex, thus easy to optimize.
\\
\hline
Hinge Loss
&
$
\sum_{i,l} [1-{F_\theta}_{li}  Y_{li}]_+=\textbf{1}_{N}^\top [\textbf{1}_{N\times L}- F_\theta \circ Y]_+
\textbf{1}_{L}
$
where $[x]_+=max(0,x)$, and $\textbf{1}_{N}\in \mathbb{R}^{N}$ is a column
vector with all ones.
&
The hinge loss function is not smooth  but subgradient descent can be used to optimize it.
It is the most common loss
function in SVM.\\
\hline
Logistic Loss
&
$\sum_{i,l} ln[1+e^{-{F_\theta}_{li}  Y_{li}}] =\textbf{1}_{N}^\top
ln\left[\textbf{1}_{N\times L}+e^{-F_\theta \circ Y}\right]
\textbf{1}_{L}
$
&
This loss function is also smooth and convex, and is usually used in regression problem.\\
\hline
\end{tabular}
\end{table}

These loss functions introduced in Table \ref{tab:loss} have been used widely in various learning problems.
One common feature of these loss function is that a sample-wise loss function is applied to
each training sample equally and then
the losses of all the samples are summed up to obtain the final
overall loss.
The sample-wise loss functions are of exactly the same {form} with the same parameter (if they have parameters).
The basic assumption behind this loss function is that the training samples are of the same importance.
However, due to the limitation of the sampling technology and noises occurred during the sampling procedure,
there are some noisy and outlying samples in real-world applications.
If we use the transitional loss functions listed in   Table \ref{tab:loss},
the noisy and outlying training samples will {play} more important roles even than the
good samples.
Thus the predictors learned by minimizing the transitional loss functions are not robust to the
noisy and outlying training samples, and could bring a high error rate when applied to the
prediction of test samples.

Recently, regularized correntropy framework has been proposed for robust
pattern recognition problems \cite{He2011,lu2012heterogeneous,li2014graph,lu2013exhaustive}.
In \cite{He2012}, He et, al argued that the classical mean square
error (MSE) criterion is sensitive to outliers, and introduced the
correntropy to improve the robustness of the presentation. Moreover, the
$l_1$ regularization
scheme is imposed on the correntropy to learn robust and sparse
representations.
Inspired by their work, we propose to use the regularized correntropy as a criterion
to compare the prediction results and the true class labels.
We use correntropy to compare the predicted labels and the true labels, instead of comparing the feature of test sample and its reconstruction from the training samples in He et, al's work.
Moreover, an $l_2$ norm regularization is introduced to control the complexity of the predictor.
In this way, the predictor learned by maximizing the correntropy between prediction results and
the true labels will be robust to the noisy and outlying training samples.
The proposed classification Machine Maximizing the Regularized CorrEntropy, which is called RegMaxCEM, is supposed to be more insensitive to outlining samples than the ones with transitional loss functions.
Yang et, al. \cite{Yang2012} also proposed to use correntropy to compare predicted class labels and true labels. However, in their framework, the target is to learn the class labels of the unlabeled samples
in a transductive semi-supervised manner, while we try to learn the parameters for the class label predictor in
a supervised manner.

The rest of this paper is structured as follows: In Section \ref{sec:method}, we propose the regularized maximum correntropy machine
by constructing an objective function based on the maximum correntropy criterion (MCC) and developing an expectation -- maximization (EM)  based alternative algorithm for
its optimization. In Section \ref{sec:exp}, the
proposed methods are validated by conducting extensive
experiments on two challenging pattern classification tasks. Finally, we give the conclusion in Section \ref{sec:conclusion}.

\section{Regularized Maximum Correntropy Machine}
\label{sec:method}

In this section we will introduce the classification machine maximizing the correntropy between the predicted class labels and the true class labels,
while keeping the solution as simple as possible.

\subsection{Objective Function}

To design the predictors $f^l_\theta(x)$,
we first represent the data sample $x$ as $\widetilde{x}$ in the linear space and the kernel space as:

\begin{equation}
\label{equ:represent}
\begin{aligned}
\widetilde{x}=
\left\{\begin{matrix}
x , & (linear), \\
K(\cdot,x), & (kernel),
\end{matrix}\right.
\end{aligned}
\end{equation}
where $K(\cdot, x) = [K(x_1, x), \cdots,K(x_N, x)]^\top \in \mathbb{R}^N$ and $K(x_i,x_j)$ is a kernel function between
$x_i$ and $x_j$.
Then a linear predictor $f^l_\theta(x)$ will be designed to predict whether the sample belongs to the $l$-th class as

\begin{equation}
\begin{aligned}
f^l_\theta(x)=w_l^\top \widetilde{x} + b_l,~l=1,\cdots,L,
\end{aligned}
\end{equation}
where $\theta=\{(w_l,b_l)\}_{l=1}^L$ is the parameters of the predictors,
$w_l\in \mathbb{R}^D$ is the linear coefficient vector and  $b_l\in \mathbb{R}$ is a bias term for the $l$-th predictor.
The target of predictor designing is to find the optimal parameters to have the prediction result
$f^l_\theta(x_i)$
of the $i$-th sample to fit its true class label indicator
$Y_{li}$ as well as possible,
while keeping the solution as simple as possible.
To this end, we consider the following two problems simultaneously when designing the objective function:

\begin{description}
\item[Prediction Accuracy Criterion based on Correntropy]
To consider the prediction accuracy,
we could learn the predictor parameters by
minimizing a loss function listed in Table \ref{tab:loss} as

\begin{equation}
\begin{aligned}
\underset{\theta}{min}
L(F_\theta,Y)
\end{aligned}
\end{equation}

However, as we mentioned in Section \ref{sec:intro}, all these loss functions are applied to all the training samples equally, which is not robust to the noisy samples and outlying samples.
To handle this problem, instead of minimizing a loss function to learn the predictor, we use the MCC \cite{He2011} framework to learn the predictor by maximizing the correntropy between the predicted results and the true labels.

{\textbf{Remark 1}: In previous studies, it has been claimed that the MCC is insensitive to outliers. For example, in \cite{He2011}, it is claimed that ``the maximum correntropy criterion, ... is much more
insensitive to outliers." Based on this fact, we assume that the predictors developed based on MCC should also be insensitive to outliers.}

Correntropy  is a
generalized similarity measure
between two arbitrary random variables $A$ and $B$.
However,
the joint probability density function of $A$ and $B$  is usually unknown, and
only a finite number of samples of them are available as $\{(a_i , b_i )\}^d_{i=1}$. It leads to the following sample estimator of correntropy:

\begin{equation}
\begin{aligned}
V(A, B) = \frac{1}{d}
\sum_{i=1}^d
g_\sigma(a_i - b_i ),
\end{aligned}
\end{equation}
where $g_\sigma(a_i - b_i )=exp\left ( -\frac{(a_i-b_i )^2}{2\sigma^2} \right )$ is a Gaussian kernel function, and $\sigma$ is a kernel width parameter. For a learning system, MCC is defined as

\begin{equation}
\label{equ:gkernel}
\begin{aligned}
max_\vartheta \frac{1}{d}
\sum_{i=1}^d
g_\sigma(a_i - b_i )
\end{aligned}
\end{equation}
where $\vartheta$ is the parameter to be optimized in the criterion so that $B$ is as correlated to $A$ as possible.

{\textbf{Remark 2}: $\vartheta$ is usually a parameter to define $B$, but not the kernel function parameter $\sigma$. In the learning system, we try to learn $\vartheta$ so that with the learned $\vartheta$, $B$ is correlated to $A$. For example, in this case, $A$ is the true class label matrix while $B$ is the predicted class label matrix, and $\vartheta$ is the predictor parameter to define $B$.}

To adapt the MCC framework to the predictor learning problem,
we let  $A$ be the prediction result matrix $F_\theta$
parameterized by $\theta$, and  $B$ be the true class label matrix $Y$, and we want to find the predictor parameter $\theta$
such that $F_\theta$ becomes as
correlated to $Y$ as possible under the MCC framework.
Then, the following correntropy-based
predictor learning model will be obtained:

\begin{equation}
\label{equ:MCC}
\begin{aligned}
\underset{\theta}{max} &V(F_\theta,Y),\\
&V(F_\theta,Y)=\frac{1}{L\times N} \sum_{l=1}^L \sum_{i=1}^N g_\sigma({F_\theta}_{li} - Y_{li})
\end{aligned}
\end{equation}
Please notice that in \cite{He2011}, MCC is used to measure the similarity between a
test sample and its
sparse linear representation of training samples, while in this work it is used to measure the
similarity between the predicted class label and its true label.
{Also note that the dependence on $\sigma$ in (\ref{equ:MCC}) and later (\ref{equ:object}), (\ref{equ:P}) relies on the dependence of the kernel function $g_\sigma(\cdot)$.
In our experiments, the $\sigma$ value is calculated as $\sigma=\frac{1}{2\times L\times N} \sum_{l=1}^L \sum_{i=1}^N \|{F_\theta}_{li} - Y_{li}\|_2^2$ following \cite{He2011}.}

\item[Predictor Regularization]
To control the complexity of the $l$-th predictor independently, we introduce
the $l_2$-based regularizer $||w_l||^2$ to the coefficient vector $w_l$ of the $l$-th predictor. We assume that the predictors of different classes are equally important,
and the following regularizer is introduced for multi-class learning problem:

\begin{equation}
\label{equ:regul}
\begin{aligned}
\underset{\{w_l\}_{l=1}^L}{min}
\frac{1}{L}\sum_{l=1}^L
||w_l||^2
\end{aligned}
\end{equation}

{\textbf{Remark 3}:The $l_2$ norm is also used by support vector regression as a measure of model complexity. However, in support vector classification, this regularization term is either obtained by a ``maximal margin" regularization or obtained by a ``maximal robustness" regularization for certain type of feature noises \cite{xu2009robustness}. Thus  our $l_2$ norm regularization term can also be regarded as a term to seek maximal margin or robustness.}

{\textbf{Remark 4}: The $l_2$-regularization is used in comparison to the $l_1$-regularization
in our model. Using $l_1$-regularization we can seek the sparsity of the predictor coefficient vector, but it cannot guarantee the minimal model complexity, maximal margin or maximal robustness like the $l_2$-regularization, thus we choose to use the $l_2$-regularization. In the future, we will explore the usage of $l_1$-regularization to see if the prediction results can be improved.}

\end{description}

By substituting $\theta=\{w_l,b_l\}_{l=1}^L$, ${F_\theta}_{li}=f^l_{w_l,b_l}(x_i)$, and combining both the
predictor regularization term in (\ref{equ:regul})
and the prediction accuracy criterion term based on correntropy in (\ref{equ:MCC}),
we obtain the following maximization problem for the maximum correntropy machine:

\begin{equation}
\label{equ:object}
\begin{aligned}
\underset{\{(w_l,b_l)\}_{l=1}^L}{max} &
\frac{1}{L\times N} \sum_{l=1}^L \sum_{i=1}^N g_\sigma(f^l_{w_l,b_l}(x_i) - Y_{li})
-\alpha \frac{1}{L} \sum_{l=1}^L
||w_l||^2
\end{aligned}
\end{equation}
where $\alpha$ is a tradeoff parameter.
This optimization problem is based on correntropy using a Gaussian kernel function $g_\sigma(x)$.
It treats the prediction of individual training samples of individual classes differently.
By this way, we can give more emphasis on samples with correctly predicted class labels,
while
those noisy or outlying training samples  will
have small contributions to the correntropy.
In fact, when the regularizer term is introduced, (\ref{equ:object}) is  a case of the regularized correntropy framework \cite{He2012}.

\subsection{Optimization}

Due to the nonlinear attribute of the kernel function $g_\sigma(x)$ in the objective function in (\ref{equ:object}),
direct optimization is difficult.
{An attribute of the kernel function $g_\sigma(x)$ is that its derivative is also the same kernel function, and if we set its derivative to zero to seek the optimization of the objective, it is not easy to obtain a close form solution.}
However, according to the property of the convex conjugate function, we have:

\begin{description}
\item[Proposition 1]
There exists a convex conjugate function $\varphi$ of
$g_\sigma(x)$ such that

\begin{equation}
\label{equ:conjugate}
\begin{aligned}
g_\sigma(x)=max_{p} (p ||x||^2 - \varphi(p))
\end{aligned}
\end{equation}
and for a fixed $x$, the maximum is reached at $p=-g_\sigma(x)$.
{This Proposition is taken from \cite{yuan2009robust},
which is further derived from the theory of convex conjugated functions.
It is further discussed and used in many applications such as \cite{He2011,He2012,lu2012image,lu2011kernel}.}
\end{description}

By substituting (\ref{equ:conjugate}) to (\ref{equ:object}),
we have the augmented
optimization problem in an enlarged parameter space

\begin{equation}
\label{equ:enlarged}
\begin{aligned}
\underset{\{(w_l,b_l)\}_{l=1}^L, P}{max}
&
\frac{1}{L\times N} \sum_{l=1}^L \sum_{i=1}^N
\left [
P_{li} ||f^l_{w_l,b_l}(x_i) - Y_{li}||^2 - \varphi(P_{li})
\right ]
-\alpha \frac{1}{L} \sum_{l=1}^L
||w_l||^2
\\
&=
\frac{1}{L\times N} \sum_{l=1}^L \sum_{i=1}^N
\left [
P_{li} ||w_l^\top \widetilde{x}_i+b_l
- Y_{li}||^2 - \varphi(P_{li})
\right ]
-\alpha \frac{1}{L} \sum_{l=1}^L
||w_l||^2,
\end{aligned}
\end{equation}
where $P=[P_{li}]\in \mathbb{R}^{N\times L}$ are the auxiliary variable matrix.
To optimize (\ref{equ:enlarged}), we adapt the
EM framework to solve $P$ and $\{(w_l,b_l)\}_{l=1}^L$
alternately.

\subsubsection{Expectation Step}

In the expectation step of the EM algorithm, we calculated the auxiliary variable matrix $P$
by fixing $\theta$. Obviously, according to \textbf{Proposition 1},
the maximum of (\ref{equ:enlarged}) can be reached at

\begin{equation}
\label{equ:P}
\begin{aligned}
P&=-g_\sigma(F_\theta - Y),\\
P_{li}&=-g_\sigma(w_l^\top \widetilde{x}_i+b_l
- Y_{li}).
\end{aligned}
\end{equation}
Note that $g_\sigma(X)$ is the  element-wise Gaussian function. With fixed predictor parameters, the
auxiliary variable $-P_{li}$ can be regarded as confidence of prediction result of the $i$-th training sample regarding to the $l$-th class.
The better the $l$-th prediction result of the $i$-th sample fits the true label $Y_{li}$, the larger
the $-P_{li}$ will be.

{\textbf{Remark 5}:
It is interesting to see if there is any relation between the auxiliary variables in $P$ and the slack variables in SVM.
Actually, both the auxiliary variables in $P$ and the slack variables in SVM can be viewed as measures of classification losses. The slack variables in SVM are the upper boundaries of hinge losses of the training samples, while the auxiliary variables in $P$ are a dissimilarity measure between the predicted labels and the true labels under the framework of the MCC rule, which is also a loss function. Meanwhile, the auxiliary variables in $P$ also play a role of weights of different training samples as in (\ref{equ:enlarged}), so that the learning can be robust to the noisy labels, but the auxiliary variables in SVM do not have such functions.}

{\textbf{Remark 6}: In the expectation step, we actually solve an alternative optimization of solving $P$ while fixing $\{(w_l,b_l)\}_{l=1}^L$. However, according to \textbf{Proposition 1}, the solution for this optimization problem is in the form of (\ref{equ:P}), which can be calculated directly and makes it an expectation step of the EM algorithm.}

\subsubsection{Maximization Step}

In the maximization step of the EM algorithm, we solve the
predictor parameters $\{(w_l,b_l)\}_{l=1}^L$ while fixing $P$.
The optimization problem in (\ref{equ:enlarged}) turns to

\begin{equation}
\label{equ:Mobjective}
\begin{aligned}
\underset{\{(w_l,b_l)\}_{l=1}^L}{max}
&\frac{1}{L\times N} \sum_{l=1}^L \sum_{i=1}^N
\left [
P_{li} ||w_l^\top \widetilde{x}_i+b_l
- Y_{li}||^2 - \varphi(P_{li})
\right ]
-\alpha \frac{1}{L} \sum_{l=1}^L
||w_l||^2.
\end{aligned}
\end{equation}
Noticing $P_{li}<0$ and removing  terms irrelevant to $w_l$ and $b_l$, the maximization problem in (\ref{equ:Mobjective}) can be reformulated as the
following dual minimization problem:

\begin{equation}
\label{equ:Daul}
\begin{aligned}
\underset{\{(w_l,b_l)\}_{l=1}^L}{min} O(w_1,b_1,\cdots,w_L,b_L)&,
\\
O(w_1,b_1,\cdots,w_L,b_L)=&
\frac{1}{L\times N}
\sum_{l=1}^L
\sum_{i=1}^N
(
-P_{li} ||w_l^\top \widetilde{x}_i+b_l
- Y_{li}||^2
)
+\alpha \sum_{l=1}^L
||w_l||^2.
\end{aligned}
\end{equation}
To simplify the {notations}, we define a vector $u_l=[u_{l1},\cdots,u_{lN}]^\top\in \mathbb{R}^N$ so that
$u_{li}^2=-\frac{1}{N} P_{li}$.
With $u_l$, the objective function in (\ref{equ:Daul}) can be rewritten as

\begin{equation}
\begin{aligned}
O(w_1,b_1,\cdots,w_L,b_L)=&
\frac{1}{L} \sum_{l=1}^L
\left [
||u_{li}(w_l^\top \widetilde{x}_i+b_l
- Y_{li})||^2
+\alpha
||w_l||^2
\right ]\\
=&
\frac{1}{L} \sum_{l=1}^L
\left [
(w_l^\top
\overline{X}_l +b_l u_l^\top - \overline{Y}_l)
(w_l^\top
\overline{X}_l +b_l u_l^\top - \overline{Y}_l)^\top
+\alpha
w_l^\top w_l
\right ],
\end{aligned}
\end{equation}
where $\overline{X}_l=[u_{l1}\widetilde{x}_1,\cdots,u_{lN}\widetilde{x}_N]\in \mathbb{R}^{D\times N}$
is the matrix containing all the training sample feature vectors weighted by $u_l$, and $\overline{Y}_l=[u_{l1} Y_{l1},\cdots, u_{lN} Y_{lN}]\in \mathbb{R}^{N}$ is the $l$-th row of $Y$ weighted by $u_l$.

Obviously, the optimization problem in (\ref{equ:Daul}) is a linear least squares problem.
Analytical solution for Problem (\ref{equ:Daul}) could be obtained easily.
By setting the derivative of $O(w_1,b_1,\cdots,w_L,b_L)$ with regard to $b_l$ to zero, we have

\begin{equation}
\label{equ:bl}
\begin{aligned}
&\frac{\partial O(w_1,b_1,\cdots,w_L,b_L)}{\partial b_l}=\frac{1}{2L}
(w_l^\top
\overline{X}_l +b_l u_l^\top - \overline{Y}_l) \textbf{1}_N =0\\
&\Rightarrow b_l= \frac{(\overline{Y}_l - w_l^\top
\overline{X}_l ) \textbf{1}_N}{u_l^\top \textbf{1}_N}
=\overline{y}_l - w_l^\top \overline{x}_l,
\end{aligned}
\end{equation}
where $\overline{y}_l=\frac{\overline{Y}_l \textbf{1}_N}{u_l^\top \textbf{1}_N}$
and $\overline{x}_l =\frac{  \overline{X}_l \textbf{1}_N} {u_l^\top \textbf{1}_N}$.
By substituting (\ref{equ:bl}) to $O(w_1,b_1,\cdots,w_L,b_L)$, we have

\begin{equation}
\begin{aligned}
O(w_1,\cdots,w_L)=&
\frac{1}{L} \sum_{l=1}^L
\left \{
[w_l^\top
(\overline{X}_l-\overline{x}_l  u_l^\top) - (\overline{Y}_l- \overline{y}_l  u_l^\top  )]
[w_l^\top
(\overline{X}_l-\overline{x}_l  u_l^\top) - (\overline{Y}_l- \overline{y}_l  u_l^\top  )]^\top
+\alpha
w_l^\top w_l
\right \}
\end{aligned}
\end{equation}
By setting the derivative of $O(w_1,\cdots,w_L)$ with regard to $w_l$ to zero, we have
the optimal solution $w_l^*$

\begin{equation}
\label{equ:wl}
\begin{aligned}
&\frac{\partial O(w_1,\cdots,w_L)}{\partial w_l}=\frac{1}{2L}
[2
(\overline{X}_l-\overline{x}_l  u_l^\top)
(\overline{X}_l-\overline{x}_l  u_l^\top)^\top
w_l
- 2 (\overline{X}_l-\overline{x}_l  u_l^\top)(\overline{Y}_l- \overline{y}_l  u_l^\top  )^\top
+2 \alpha w_l
] =0\\
&\Rightarrow w_l^*= [(\overline{X}_l-\overline{x}_l  u_l^\top)
(\overline{X}_l-\overline{x}_l  u_l^\top)^\top + \alpha I]^{-1}
(\overline{X}_l-\overline{x}_l  u_l^\top)(\overline{Y}_l- \overline{y}_l  u_l^\top  )^\top,
\end{aligned}
\end{equation}
where $I$ is an ${D\times D}$ identity matrix.
Then we substitute $w_l^*$ to (\ref{equ:bl}),  and we will have
the optimal solution of $b_l^*$,

\begin{equation}
\label{equ:blop}
\begin{aligned}
b_l^*=\overline{y}_l - {w_l^*}^\top \overline{x}_l
\end{aligned}
\end{equation}

\subsection{Algorithm}

Algorithm 1 summarizes the predictor parameter  learning procedure
of RegMaxCEM.
The E-step and the M-step will be repeated for $T$ times.

\begin{algorithm}[!h]
\caption{RegMaxCEM Learning Algorithm.}\label{alg:AdapGrNMFMultiK}
\begin{algorithmic}
\STATE \textbf{Input}: Training set: $\mathcal{D}=\{(x_i,y_i)\}_{i=1}^N$;

\STATE Initialize the auxiliary variable matrix $P^{0}=-\textbf{1}_{L\times N}$;

\STATE Represent each sample $x_i$ as $\widetilde{x}_i$ as in (\ref{equ:represent});

\FOR{$t=1,\cdots,T$}

\STATE \textbf{Maximization-Step}: Update the predictor parameters $\theta^t=\{(w_l^t,b_l^t)\}_{l=1}^L$ as in (\ref{equ:wl})
and (\ref{equ:blop}) by fixing $P^{t-1}$.

\STATE \textbf{ Expectation-Step}: Update the auxiliary variable matrix  $P^{t}$ as in (\ref{equ:P}) by fixing the predictor parameters $\theta^t$.

\ENDFOR

\STATE \textbf{Output}: Predictor parameters $\theta^T=\{(w_l^T,b_l^T)\}_{l=1}^L$.

\end{algorithmic}
\end{algorithm}

\section{Experiments}
\label{sec:exp}

In the experiments, we will evaluate the proposed classification method
on two challenging pattern classification tasks ---
bacteria identification \cite{Pei2012} and
prediction of DNA-binding sites in proteins \cite{DNAbinding2013}.


\subsection{Experiment I: Bacteria Identification}

\subsubsection{Dataset and Setup}

High-precision identification of bacteria is quite important for the
diagnosis of cancers and bacterial infections.
Recently, ensemble
aptamers (ENSaptamers), which utilizes  a small set of nonspecific DNA sequences,
has been proposed to provide an effective platform for the
detection of bacteria \cite{Pei2012}.
ENSaptamers is
a sensor array with seven sensors, and each sensor
is designed  using a DNA
element.

For the experiment, we collected in total 66 samples of
6 different bacteria, including
S.tyohimurium, S.flexneri, E.coli (CAU 0111), S.sonnei, S.typhi and E.coli (ATCC 25922).
The number of samples for each bacteria varies from 9 to 13.
Given an unknown bacteria sample with its fluorescence response patterns of ENSaptamer,
the task is to identify which bacteria it is.
To this end
the seven fluorescence response patterns of ENSaptamer against the sample will be
used to construct the 7-dimensional feature vector, and then the sample will be classified into one of the
known bacteria using the RegMaxCEM predictor.

To conduct the experiment, we randomly split the entire dataset into two non-overlapping subsets --- the training set and the test set.
33 samples were used as training sample in the training set, while
the remaining 33 ones as test samples.
The predictor parameters of RegMaxCEM were trained using the feature vectors and
class labels of the training samples.
Then the class labels of the test samples were predicted by the trained predictor,
and compared to their true labels to calculate the classification accuracy.
The random split process (training/test) was repeated for ten times and the
accuracies over these ten splits were reported  as classification performance.

\subsubsection{Results}

\begin{figure}[htbp!]
\centering
\subfigure[Linear representation]{
\includegraphics[width=0.8\textwidth]{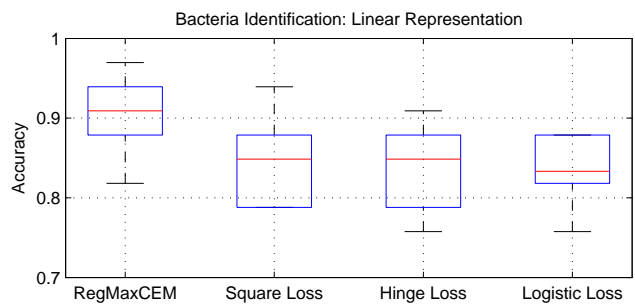}}
\subfigure[Kernel representation]{
\includegraphics[width=0.8\textwidth]{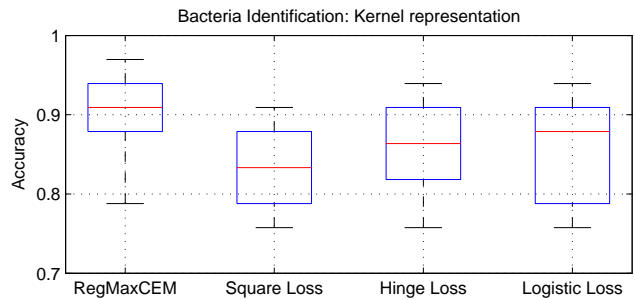}}
\caption{Boxplots of accuracies of bacteria identification.}
\label{fig:FigBact}
\end{figure}

We compare our proposed method against other loss function based classifiers, including
square loss, hinge loss and logistic loss.
0-1 loss is the simplest loss function, but difficult to optimize, thus is not compared in the
experiment.
The boxplots of accuracies of different methods using both linear and kernel representations are
illuminated in Figure \ref{fig:FigBact}.
As shown in  Figure \ref{fig:FigBact}, predictor produced by
maximizing the correntropy yields improvements over other loss functions.
Given the extremely small
variation of classification accuracies over the ten splits,
though the improvement of the accuracies are not
large in absolute terms (around 0.1), it is  consistent and significant.
To verify whether the improvements are statistically significant,
we performed the paired t-tests to the accuracies of the proposed method and other compared methods.
The null hypothesis of the T-test is that the accuracies of the proposed method and the compared methods
come from distributions with     equal means.
The $P$ values of the t-tests are reported as measurements of statistically significance.
A low $P$ value implies that the difference between the proposed method and the compared methods are statistically  significant.
The $P$ values are reported in Table \ref{tab:TabBact}.
As we can see from the table, all the improvements archived by RegMaxCEM, for both linear representation and
kernel representation, are statistically  significant at the 0.05 significance level.
This is not surprising: There are some noisy and outlying samples in the training set, which have been utilized
by the methods with square loss, hinge loss or logistic
loss as equally as other samples, thus they bring some bias to the predictor.
However, the RegMaxCEM  has the potential of
filtering these samples,
which can result
in reliable learning of predictors in practice.
It is also interesting to notice that the square loss, hinge loss and logistic
loss have archived very similar classification accuracies.
Though they used different loss functions, these loss functions are applied to the training samples
equally.

\begin{table}[htbp!]
\centering
\caption{$P$ values of  paired T-tests on accuracies of ten splits
of RegMaxCEM and compared methods on bacteria identification.}
\label{tab:TabBact}
\begin{tabular}{|l|c|}
\hline
\multicolumn{2}{|c|}{Linear representation}\\
\hline
\hline
Compared methods & $P$ values \\
\hline
\hline
Square Loss& 0.0266\\
\hline
Hinge Loss& 0.0243\\
\hline
Logistic Loss& 0.0115\\
\hline
\end{tabular}
\begin{tabular}{|l|c|}
\hline
\multicolumn{2}{|c|}{Kernel representation}\\
\hline
\hline
Compared methods & $P$ values \\
\hline
\hline
Square Loss& 0.0118\\
\hline
Hinge Loss& 0.0224\\
\hline
Logistic Loss& 0.0095\\
\hline
\end{tabular}
\end{table}

\subsection{Experiment II: DNA-Binding Site Prediction}

It is very important to predict the DNA-binding sites in proteins
for understanding the
molecular mechanisms of protein-DNA interaction.
In this experiment, we will evaluate the proposed method for prediction of DNA-binding sites \cite{DNAbinding2013}.

\subsubsection{Dataset and Setup}

The PDNA-62 database for DNA-binding site prediction has been used in this experiment.
This database contains 8,163 sites in proteins in total.
Among these sites, 1,215 of them are DNA-binding sites, while the remaining 6,948 sites are non-binding sites.
We select 1,000 DNA-binding sites and 5,000 non-binding sites
from the PDNA-62 database to construct our database for the experiment.
Given a candidate site, the goal of DNA-binding site prediction is to predict whether it is a
DNA-binding site or not.
To this end, the evolutionary information, solvent accessible surface area and the protein
backbone structure
features were extracted from the site, and then combined to construct the feature vector.
The feature vector was further inputted into the classifier to distinguish
DNA-binding sites from the non-binding sites \cite{DNAbinding2013}.

To conduct the experiment, we employed the 10-fold cross validation.
The database was split into 10 non-overlapping folds randomly, one of which was used as the test set, while the rest 9 of them were used as the training set. The procedure was repeated for 10 times so that each fold was used as the test set once.

The prediction performance was  measured by the receiver operating characteristic (ROC) and recall-precision curves.
{The usage of ROC curve is mainly due to the imbalanced classes.}
The ROC curve is created by plotting false positive rate (FPR) against true positive rate (TPR), while
recall-precision curve is obtained by ploting recall against precision.
The FPR, TPR, recall and precision are defined as:

\begin{equation}
\begin{aligned}
&FPR=\frac{FP}{FP+TN},~TPR=\frac{TP}{TP+FN},\\
&recall=\frac{TP}{TP+FN},~precision=\frac{TP}{TP+FP},
\end{aligned}
\end{equation}
where $TP$ is the number of DNA-binding sites predicted correctly,
$FP$ is the number of non-binding sites predicted as DNA-binding sites wrongly,
$TN$ is the number of non-binding sites predicted correctly,
while $FN$ is the number of DNA-binding sites predicted as non-binding sites wrongly.
For a better predictor, its ROC curve should be closer to the  top left corner of the figure,
while the recall-precision curve should be  closer to the top right corner.
Besides the two curves, area under the ROC curve (AUC) is also used as a single measurement of the prediction.
A better predictor will have a larger  AUC value.

\subsubsection{Results}

\begin{figure}[htbp!]
\centering
\subfigure[ROC of linear presentation]{
\includegraphics[width=0.45\textwidth]{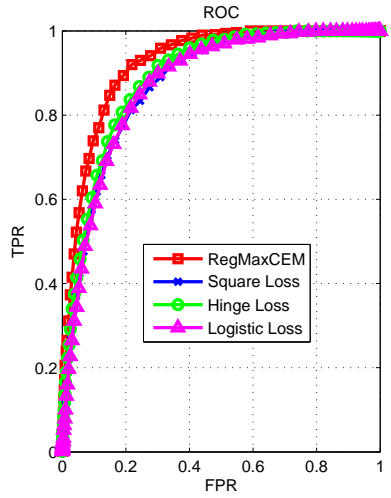}}
\subfigure[Recall-precision curve of linear presentation]{
\includegraphics[width=0.45\textwidth]{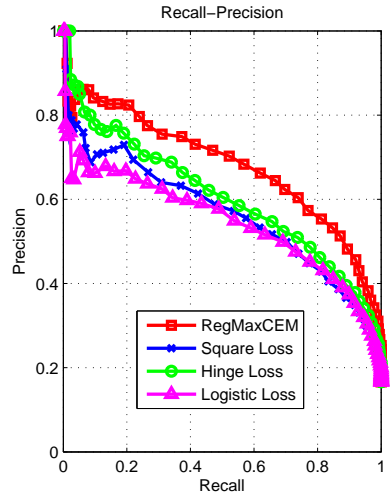}}
\subfigure[ROC of kernel presentation]{
\includegraphics[width=0.45\textwidth]{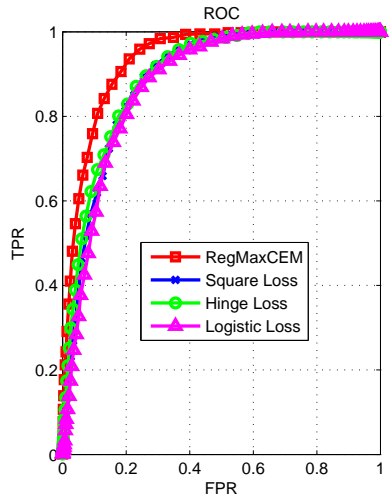}}
\subfigure[Recall-precision curve of kernel presentation]{
\includegraphics[width=0.45\textwidth]{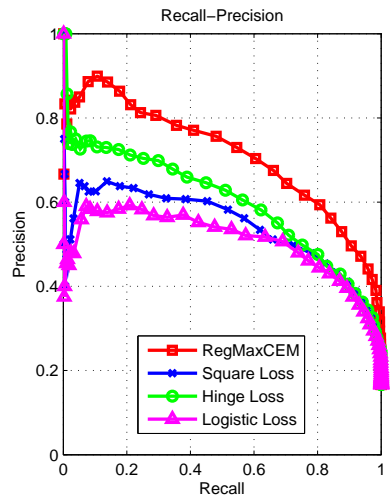}}
\caption{ROC and recall-precision curves on  DNA-Binding site prediction experiment using
both linear and kernel representations.}
\label{fig:FigDNAB}
\end{figure}

The ROC and recall-precision curves of the proposed method and compared methods are
reported in Figure \ref{fig:FigDNAB}.
The predictors using linear and kernel representations are both illuminated.
The AUC values of the ROC curves are reported in Table \ref{tab:TabDNAB} as well.
Overall the proposed methods clearly outperform the other
methods significantly,
although there is some
variability in prediction performance over different representation types.
From Table \ref{tab:TabDNAB}, we could see that the accuracy of the predictor
is slightly increased by using the kernel representation
instead of the linear representation. The regularized correntropy based predictors gives much better results than other methods on both representations. An interesting result from the DNA-binding prediction
on this dataset is that the predictor with the  hinge loss function outperforms other two methods.

\begin{table}[htbp!]
\centering
\caption{AUC values of ROC curves on DNA-Binding site prediction experiment.}
\label{tab:TabDNAB}
\begin{tabular}{|l|c|}
\hline
\multicolumn{2}{|c|}{Linear representation}\\
\hline
\hline
Mehtods & AUC \\
\hline
\hline
RegMaxCEM &0.9226 \\
\hline
Square Loss &0.8768 \\
\hline
Hinge Loss &0.8908 \\
\hline
Logistic Loss &0.8747 \\
\hline
\end{tabular}
\begin{tabular}{|l|c|}
\hline
\multicolumn{2}{|c|}{Kernel representation}\\
\hline
\hline
Mehtods & AUC \\
\hline
\hline
RegMaxCEM &0.9344 \\
\hline
Square Loss &0.8891 \\
\hline
Hinge Loss &0.8961 \\
\hline
Logistic Loss &0.8776 \\
\hline
\end{tabular}
\end{table}

\section{Conclusion and Future Work}
\label{sec:conclusion}

In this paper, we present a novel regularized predictor learning model
for multi-class pattern recognition problems.
The predictor is learned by maximizing the correntropy between the prediction results and the true class labels.
By applying the MCC rule, we could treat different training samples differently, so that the noisy and
outlying training samples have less impact on the learning of predictors.
Compared with the existing predictor models with various loss functions,
it is robust to the noisy and
outlying training samples.
The experiments
on bacteria identification and DNA-binding site prediction show that a good
predictor may benefit much from a well designed loss function based on MCC.
The proposed method outperformed the predictor with other popularly used loss functions.
In the future, we will investigate if the regularized maximum correntropy framework can be used to regularize ranking score learning \cite{wang2014sparse,lu2009generalized}, data representation \cite{sun2015non,lu2006unsupervised,lu2008unsupervised,lu2011spectral,lu2013latent,lu2011latent,lu2005regularized,zhao2010action,lu2007unsupervised}
Moreover, we also plan to extend the proposed regularized correntropy based classifier for wireless sensor network \cite{qingquan2010context,sunprimate,wu2013fast,sun2013multi,sun2012unsupervised,hu2012neuro,sun2011mobile}, computer vision \cite{lu2011spatial,lu2011contextual,lu2010gaussian,lu2005iterative,lu2006publishing,lu2008semi,lu2008semi,lu2009context,lu2009image}, and computer network security \cite{InTrust14-posture,6587320,Xu:2014:AED:2578044.2555613,DBLP:conf/codaspy/XuZXY13,CNS:Evasion,5887351,xu2010trustworthy}.

\end{document}